\documentclass[letterpaper]{article} 
\usepackage{aaai25}  
\usepackage{times}  
\usepackage{helvet}  
\usepackage{courier}  
\usepackage[hyphens]{url}  
\usepackage{graphicx} 
\usepackage{natbib}  
\usepackage{caption} 
\usepackage{algorithm}
\usepackage{algorithmic}

\usepackage{newfloat}
\usepackage{listings}
\DeclareCaptionStyle{ruled}{labelfont=normalfont,labelsep=colon,strut=off} 
\floatstyle{ruled}
\newfloat{listing}{tb}{lst}{}
\floatname{listing}{Listing}
\title{QGen Studio: An Adaptive Question-Answer Generation, Training and Evaluation Platform}
\author {
    Movina Moses\equalcontrib\textsuperscript{\rm 1},
    Mohab Elkaref\equalcontrib\textsuperscript{\rm 2},
    James Barry\textsuperscript{\rm 2},
    Shinnosuke Tanaka\textsuperscript{\rm 2},
    Vishnudev Kuruvanthodi\textsuperscript{\rm 2},
    Nathan Herr\textsuperscript{\rm 3}\footnote{Work done while at IBM.},
    Campbell D Watson\textsuperscript{\rm 1},
    Geeth De Mel\textsuperscript{\rm 2}
}
\affiliations {
    \textsuperscript{\rm 1}IBM Research\\
    \textsuperscript{\rm 2}IBM Research Europe\\
    \textsuperscript{\rm 3}University College London\\
    \{movina.moses, mohab.elkaref, james.barry, shinnosuke.tanaka, vishnudev.k\}@ibm.com, uceenhe@ucl.ac.uk, cwatson@us.ibm.com, geeth.demel@uk.ibm.com}

\begin{document}

\maketitle

\begin{abstract}

We present QGen Studio: an adaptive question-answer generation, training, and evaluation platform. QGen Studio enables users to leverage large language models (LLMs) to create custom question-answer datasets and fine-tune models on this synthetic data. It features a dataset viewer and model explorer to streamline this process. The dataset viewer provides key metrics and visualizes the context from which the QA pairs are generated, offering insights into data quality. The model explorer supports model comparison, allowing users to contrast the performance of their trained LLMs against other models, supporting performance benchmarking and refinement. QGen Studio delivers an interactive, end-to-end solution for generating QA datasets and training scalable, domain-adaptable models. The studio will be open-sourced soon, allowing users to deploy it locally.

\end{abstract}
\vspace{-1em}

%

\section{Introduction}
Large language models (LLMs) have shown remarkable capabilities in text generation, enabling the creation of large synthetic datasets with minimal human input. This offers significant potential for tasks like question-answering (QA) \cite{duan-etal-2017-question}, where access to high-quality data is crucial for training models. However, niche domains often lack tailored datasets, causing LLMs to perform poorly, as generic datasets may not align with specialized tasks \cite{ling2024domainspecializationkeymake}.

While LLMs can help mitigate this issue by generating domain-specific datasets \cite{wei2022finetunedlanguagemodelszeroshot}, the quality and relevance of these datasets can vary, complicating their alignment for specialized tasks \cite{gudibande2023falsepromiseimitatingproprietary}. The lack of standardized evaluation metrics further challenges their effectiveness \cite{nema-khapra-2018-towards}. There is a need for better methods to evaluate and ensure these datasets tailor to requirements.

To address these challenges, we present QGen Studio, an adaptive platform for question-answer generation, training, and evaluation. Users can upload documents and interactively generate QA pairs using LLMs from OpenAI, IBM watsonx, and HuggingFace using Apple's MLX framework. The platform provides an interactive interface that allows users to input example prompts while viewing the source text, giving them more control over the QA generation process. Generated QA pairs are visualized using a dataset viewer, highlighting metrics and contextualizing the data for better analysis.

QGen Studio also enables users to train models using the generated datasets and compare the performance of these custom models against existing ones. The studio supports a streamlined, end-to-end solution for synthetic QA dataset creation, visualization, and model refinement by addressing limitations of existing approaches and allowing users to build scalable, domain-adaptable QA models.

\begin{figure}[t]
\centering
\includegraphics[width=0.45\textwidth]{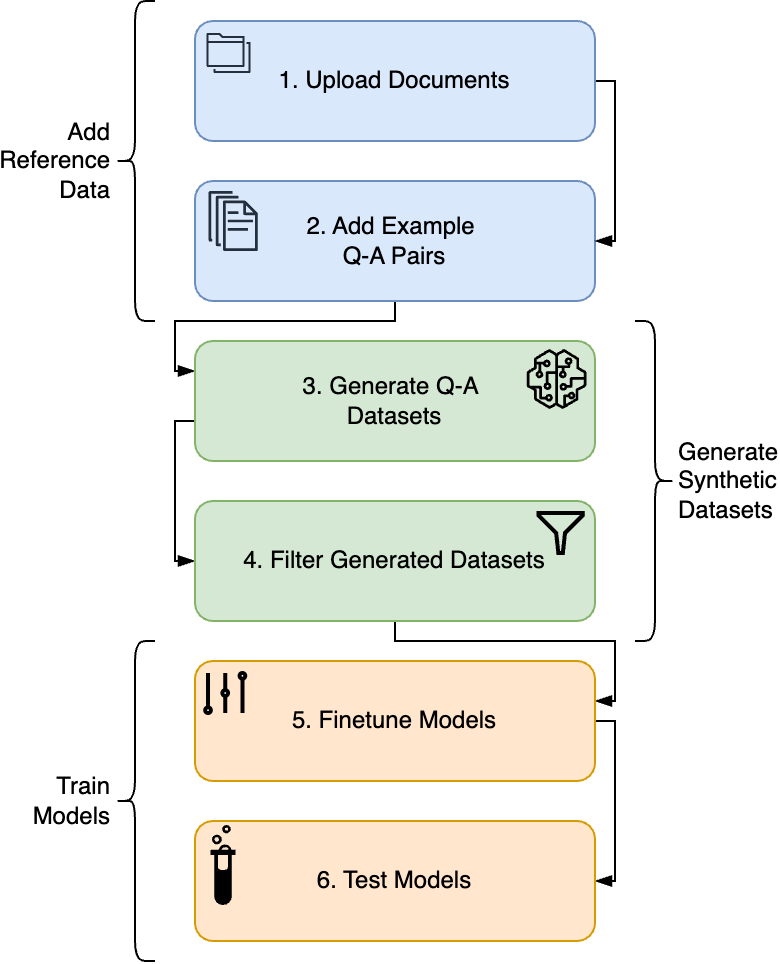}
\caption{Overview of QGen Studio}
\end{figure}

\section{Related Work}
The use of LLMs for generating synthetic data has been widely explored, demonstrating their potential in creating and augmenting datasets for NLP tasks such as question-answering (QA) \cite{long-etal-2024-llms, NEURIPS2020_1457c0d6}. Methods using rule-based generation \cite{khullar-etal-2018-automatic}, pre-trained models \cite{dong2019unifiedlanguagemodelpretraining}, and adversarial training \cite{rao-daume-iii-2019-answer} have been explored, but they often depend on existing datasets and struggle to adapt to niche domains with limited data emphasizing the need for more flexible, domain-adaptive QA generation methods. Additionally, the quality and relevance of these generated datasets can be inconsistent \cite{yu2023largelanguagemodelattributed}. 

While several platforms and tools have been developed to support dataset creation and model training \cite{patel-etal-2024-datadreamer, yu-etal-2024-localrqa}, they often limit user control over the generation process, which hinders effective exploration and understanding of the datasets.

QGen Studio addresses these limitations and tailors to the needs of niche domains with limited data availability where users would like more control over the generation process.

\section{System Overview}
QGen Studio follows a six-step pipeline that guides the user through the process of creating their datasets, training their models and evaluating its performance. In this section, we describe the technologies used in each step.

\paragraph{Document Upload}

The document upload section allows users to create document groups under which they can upload various context sources such as JSON files, PDFs and URLs to PDFs. PDFs and URLs are automatically converted to structured text using Docling \cite{auer2024doclingtechnicalreport}, split into headings and paragraphs and assigned unique IDs. The files are organized as documents within their respective document groups. The interface allows users to add, view and delete documents within each group.

\paragraph{Example Prompts}

The example prompts page allows users to scroll through documents within each group and inspect, add, or delete question-answer pairs. These pairs can be used as examples within few-shot prompts to guide dataset generation, providing more control over the type of QA pairs generated from a specific document.

\paragraph{Dataset Generation}

In the dataset generation step, users generate QA pairs from selected document groups using LLMs from OpenAI\footnote{https://platform.openai.com/docs/overview} and IBM's watsonx\footnote{https://www.ibm.com/watsonx}, or HuggingFace models via Apple's MLX library \cite{mlx2023}.
After selecting an LLM and document groups, users configure generation parameters like text chunking, number of questions, and evaluation metrics to use for ranking. Available metrics include computational metrics such as 
Bleu, Rouge, and Meteor, as well as TFIDF
and cosine similarity scores, all calculated between the generated QA pairs and their source context. Users can choose between zero-shot or few-shot prompts, with few-shot examples drawn from the Example Prompts page. Once configured, QA pairs are generated, evaluated, and stored by document group.

\paragraph{Dataset Viewer}

The dataset viewer enables users to view and interact with the generated datasets. Users can select a document group, choose evaluation metrics, and highlight either the question, answer, or context. The spans in the context are highlighted based on question and answer tokens, and if ``context" is chosen, the viewer highlights the most likely sentence from which the QA pair was generated. Users can filter and sort QA pairs by metrics, such as showing only pairs with Bleu-2 scores above 0.8. The viewer lets users visualize QA pairs for each document. If desired, they can adjust parameters, metrics, or examples to refine the generation process.

\paragraph{Model Training}

The model training section enables users to fine-tune models to better adapt them to the generated datasets. This process utilizes MLX to fine-tune an LLM with low-rank adaptation (LoRA) \cite{hu2021loralowrankadaptationlarge}. Currently, this feature is limited to Apple Silicon users, but others can download the generated datasets and train models independently. Users can choose how the dataset is processed, including options to include context, shuffle the dataset, and adjust the test and validation splits. They can modify training parameters, such as the learning rate, number of iterations, and LoRA layers. This setup allows users to utilize one model for generation and a different model for training, enhancing customization and performance for specific tasks.

\paragraph{Model Explorer}

The model explorer page allows users to compare two models side by side. Users can select a document group, document, and models they wish to evaluate. They can choose between existing models and their fine-tuned models. While viewing the document, users can pose a question related to its content, and the studio will run inference on the selected models. The explorer displays the generated responses side by side, allowing users to compare and contrast the models' performance. They can assess how the generated datasets influenced the models' by reviewing the answers in the context of the selected document. This layout helps users evaluate the strengths and weaknesses of each model, facilitating informed decisions about which model is better aligned for use.

\section{Conclusion and Future Work}
In this paper, we introduced QGen Studio, a comprehensive framework for generating and evaluating question-answer pairs using various models and custom datasets. 
This tool can be used to create QA datasets across different domains and adapt models to diverse requirements. Users can explore their datasets and fine-tuned models while visualizing performance, enabling them to make necessary adjustments in the generation and training processes.

Future work will focus on supporting additional frameworks and enabling multi-document querying for insights across diverse sources. We also plan to incorporate generation for complex question types, support multilingual generation, and extend functionality for other downstream tasks.

\bibliography{aaai25}

\end{document}